\documentclass[letterpaper]{article}
\usepackage[preprint]{aaai2027}
\usepackage[hyphens]{url}
\usepackage{graphicx}
\usepackage{natbib}
\usepackage{caption}
\usepackage{booktabs}
\usepackage{array}
\usepackage{amsmath}

\newcommand{\sourcecaption}[1]{\caption{#1 \textit{Source: Microsoft.}}}
\newcommand{\IR}{\textsc{IR}}
\newcommand{\SelIR}{Selector+\IR}
\newcommand{\SelJSON}{Selector$\rightarrow$JSON}
\newcommand{\NLgraph}{NL$\rightarrow$graph}

\title{Generating Workflow DAGs from Natural Language with Non-Reasoning LLMs}
\author{
Anand Iyer, Bhanu Khetharpal, Srinivas Upadhya, Ramkumar Rajagopal
}
\affiliations{
Microsoft\\
anandiyer@microsoft.com, bkhetharpal@microsoft.com,\\
supadhya@microsoft.com, ramkumar.rajagopal@microsoft.com
}

\begin{document}
\maketitle

\begin{abstract}
This paper addresses the problem of translating natural-language routing rules
written by business administrators into executable workflow graphs for
enterprise contact centers. Each target is an executable \textbf{workflow graph},
a directed acyclic graph
(DAG) of conditional actions with parallel branches, hit-first fallback chains,
and per-branch boolean predicates---encoded in the JSON dialect of a commercial
routing platform. We present a system for this translation and show that, with
the right neuro-symbolic decomposition, \textbf{lower-cost non-reasoning LLMs can
generate complex workflow DAGs at production-relevant quality} without expensive
extended-reasoning models. Our central diagnostic is an
\textbf{emission-density bottleneck}: on a 635-rule benchmark
(manufactured synthetic data), models select
the right graph nodes almost perfectly but drop their content---mis-wiring
attributes and boolean grouping---as the number of interdependent nodes they
must emit in a single pass grows. Acting on this diagnostic, we move
combinatorial graph construction out of the model into a
\textbf{deterministic compiler}
driven by a compact intermediate representation (\IR), with a learned
registry-selection front-end that keeps generation focused as the vocabulary
grows. Across four generation models, the full system reaches up to
$\sim$89\% LLM-judge validity and $\sim$90\% exact-match condition accuracy with
99--100\% guaranteed-valid JSON, at roughly half the per-rule prompt tokens of a
monolithic prompt. Our sharpest result, tested paired, is a
\textbf{gap-bridging effect}:
structural externalization delivers a large, significant lift where monolithic
prompting leaves the most headroom (+24 points judge validity on GPT-5.3-chat),
lifting a non-reasoning model to statistically match (by an equivalence test) a
reasoning model's out-of-the-box quality, while an $\sim$8-point frontier gap
persists. We close with a deployment path and transferable lessons for
structured-generation applications.
\end{abstract}

\section{Introduction}
\textbf{Problem and significance.}
A commercial contact-center platform routes inbound chat, voice, and SMS
conversations through queues governed by routing rules. Each rule must compile to
a \textbf{workflow definition}: a JSON DAG whose nodes are actions (raise priority,
transfer queue, notify supervisor, offer callback) and whose edges encode
conditional and hit-first execution order. Authoring this JSON by hand demands
platform expertise and is error-prone. The platform's current authoring
capability generates these workflows with a single large (``monolithic'') LLM
prompt over a \textbf{structured, templatized} rule representation; extending it
to accept \textbf{natural-language} rules would substantially widen adoption, and is the problem
this paper studies.

\textbf{Representative prompt and DAG.}
Consider the prompt: ``For customers in region US, while they wait in queue,
raise priority by 10 every 10 seconds for 3 minutes; if still unresolved at 3
minutes, end the conversation.'' Its target DAG contains a queue-wait trigger;
a region = US predicate gating the branch; raise-priority, timer, and
end-conversation action nodes; and dependency edges under which priority repeats
during the timer and the timer gates the terminal action. The target is therefore
not a flat action list but a graph specifying which predicate gates each action
and how the actions depend on one another.

The task is a constrained graph-generation problem, hard for
three reasons: \textbf{density} (production rules combine up to 15 conditional
branches across customer segments---and more actions still, since one branch can
fire several---producing large branch grids), \textbf{exactness} (one wrong value
or dropped branch changes the workflow's behavior; generated workflows are
reviewed and validated by a human before deployment, so higher first-pass
accuracy reduces correction and regeneration iterations), and \textbf{cost at
scale} (the converter runs inside an interactive authoring experience, so
per-rule token budget rules out simply deploying the most powerful reasoning
model).

\textbf{Motivation.}
Iterating on the monolithic prompt surfaced a clear insight:
\textbf{few-shot exemplars are the primary driver of accuracy}. But as the
platform grows, each new scenario
adds actions, conditions, and scenario-specific exemplars, and a single prompt
that inlines all of them on every request grows in cost and dilutes attention
over largely irrelevant context. Our architecture preserves the few-shot signal
while paying only for the vocabulary and exemplars each rule actually needs.

\textbf{Thesis.}
These constraints need not force a quality compromise. A non-reasoning model
handed a large monolithic prompt degrades on dense rules, but the reason is
specific and fixable: the model is excellent at \textbf{reading} a rule and
comparatively poor at \textbf{emitting} it as dense interdependent structure in
one pass. We name this the \textbf{emission-density bottleneck}: the model almost always selects the
correct graph nodes (node-set accuracy near-saturated) yet increasingly mis-emits
their attributes and boolean structure as node density rises, so accuracy falls
with branch-grid density, not with the difficulty of reading. The remedy is
therefore architectural rather than model-scale---move the combinatorial,
drop-prone parts of graph construction into a deterministic compiler and keep the
LLM on interpretation---letting commodity non-reasoning models produce
production-relevant DAGs and reach judge validity that is statistically
equivalent, at a pre-specified $\pm5$-point margin, to a strong reasoning model's
out-of-the-box quality.

\textbf{Contributions.}
\begin{enumerate}
\item \textbf{The emission-density bottleneck}---a diagnostic that localizes
structured-generation failure to emission, not interpretation: models select the
right graph nodes almost perfectly (node-set accuracy 98--100\% for stronger
models) yet increasingly mis-emit node attributes and boolean structure as node
density rises (Figure~\ref{fig:density}), so the fix is architectural rather than
model-scale. This generalizes as a diagnostic lens for \NLgraph\ tasks.
\item \textbf{A gap-bridging result}, stated precisely and tested paired: acting
on the diagnostic lets a non-reasoning generator statistically match
(equivalence-tested at a $\pm5$-point margin) a reasoning model's
monolithic-prompt quality on the semantic metric where monolithic prompting
leaves the most headroom (+24 points, $p<10^{-3}$), while gains for models that
already emit dense structure well are small and an $\sim$8-point frontier gap
persists---a quantified account of when structural externalization helps and when
it does not.
\item \textbf{An emerging-application study} that instantiates the
LLM-as-compiler paradigm for natural-language $\rightarrow$ workflow-DAG
generation on a real enterprise routing platform. The diagnostic drives a
concrete architecture---a compact \IR\ compiled deterministically to valid JSON,
and a learned registry selector that keeps generation focused as the vocabulary
grows---yielding validity by construction (99--100\% valid JSON; 0\%
uncompilable \IR\ across all four models) at roughly half the per-rule prompt
tokens of a monolithic prompt.
\item \textbf{A deployment path}---registry-selection as the scaling mechanism, a
planned similarity-based candidate prefilter, and a target-agnostic
compiler---plus transferable evaluation lessons for structured-generation
applications.
\end{enumerate}

\section{Task and Domain}
\textbf{The registry.}
The platform defines a fixed, versioned vocabulary---the registry---of
\textbf{event types} (triggers such as queue-wait, initial-routing, agent-response,
queue-transfer), \textbf{action types} (change-priority, transfer-to-queue, offer-callback,
send-notification, \ldots), and \textbf{condition types} with data types and operators
(numeric wait-time, arbitrary-typed context variables, boolean state predicates,
\ldots). Crucially, the registry also holds \textbf{examples}---and this is its most
important part: for every new scenario we add one or more corresponding example
JSONs (as many as the scenario's complexity warrants) alongside the vocabulary
entries, so that the learned selector can retrieve its own few-shot exemplars for
the rule at hand rather than relying on a fixed, globally-inlined example set.
The registry is the single source of truth: adding a capability (or a new
scenario's exemplars) means adding a registry entry, not changing code.

\textbf{Input / output.}
The input is a text rule; the output is workflow JSON with a trigger block (the
event), an actions block (the DAG nodes), per-action boolean predicates, and
dependency edges. Hit-first chains are encoded by chaining branches on a
skip-on-match dependency; parallel branches run independently; a residual
(``everyone else'') branch carries an empty predicate. The hard part is not the
trigger or the action set (models get those $\sim$90--100\% right) but the
\textbf{conditional structure}: which predicate gates which action, and how branches
relate.

\textbf{Benchmark.}
We evaluate on a \textbf{635-rule} benchmark spanning seven structural families
with \textbf{1--15 conditional branches per rule}---and, because a single branch can trigger
several actions, more than 15 actions (and condition$\rightarrow$action pairs) in
the densest rules---deliberately at the dense end where production rules are
hardest. Ground truth is the correct workflow for each rule. Benchmark prompts
are programmatically generated from natural-language templates: this is a
deliberate methodological choice, since templating is what lets us attach an
exact ground-truth workflow---and therefore deterministic accuracy
metrics---to every rule. Because the templates systematically vary the
condition/action values, the branch structure (segment count and order), and the
surface phrasing across the seven families, they already probe some surface
diversity; beyond this, they also inject deliberate spelling errors---present in
roughly two-thirds of prompts---and non-English (e.g., Italian) entity values,
mimicking real admin input and human inaccuracy. This directly exercises surface
robustness (noisy, multilingual phrasing over complex structure), a distinct axis
from the structural complexity above. Because our prompts couple this injected
noise with deep branch structure, we expect the harder bottleneck to remain
structural rather than surface-level; still, paraphrase robustness for workflow
generation is a known open problem \citep{xu2025}, and we do not claim to isolate
fully free-form paraphrase. (The data is proprietary; see Data Availability.)

\section{Related Work}
We connect to five bodies of work. \textbf{(a) Semantic parsing and
NL-to-structured output.} Mapping language to executable structure has a long
tradition in text-to-SQL and semantic parsing, where exact-match accuracy and
cross-schema generalization are central \citep{yu2018}. Grammar-constrained
decoding guarantees schema conformance without finetuning
\citep{geng2023,willard2023,park2025}; our compiler instead achieves validity by
construction on the back end, so it is model- and API-agnostic. \textbf{(b)
Neuro-symbolic generation}, where an LLM proposes a symbolic representation a
deterministic component executes or verifies \citep{kamali2025}; our
\IR$\rightarrow$compiler split is an instance specialized to workflow DAGs.
\textbf{(c) LLM-driven workflow and process generation}, treating the LLM as a
compiler that emits an intermediate program a runtime executes
\citep{zeng2023,fan2024,trooskens2026}. We differ on two axes: our input is text
rules (the hard NL$\rightarrow$structure step is upstream of where these systems
begin), and our back end is a total compiler---the \IR\ never fails to
compile---rather than a generate-validate-regenerate loop. Closest to us,
agentic-workflow benchmarks measure graph-structured plans with subgraph matching
and report a marked sequence-vs-graph gap even for strong models
\citep{qiao2025,xu2025,stiehle2025}; our emission-density analysis localizes why
graph quality degrades and our compiler removes the validity failure mode.
\textbf{(d) Retrieval-augmented generation and tool/registry selection}
\citep{jiang2023,mattioli2026}; our similarity-prefiltered learned selector is a
task-specific instance for registry scaling. \textbf{(e) LLM-as-judge evaluation}
and its confounds---position, verbosity, and self-enhancement biases
\citep{zheng2023,gu2024,wataoka2024}; Section~\ref{sec:setup} bounds their effect
and we report judge-independent metrics throughout. Our distinguishing
contributions are the emission-density localization, the gap-bridging result for
non-reasoning models, and a disclosed evaluation methodology.

\section{Method}
The system is a pipeline of LLM and deterministic stages, presented in the order
each stage addresses a specific measured failure. Four configurations recur in
the evaluation:
\begin{itemize}
\item \textbf{Monolithic}---one LLM call, full registry $\rightarrow$ JSON
directly (no compiler).
\item \textbf{\SelJSON}---learned registry selection, then the model emits JSON
directly.
\item \textbf{\IR-only}---\IR\ + compiler over the full registry (no selector).
\item \textbf{\SelIR}---the full system: learned selection and \IR\ + compiler.
\end{itemize}

\subsection{Monolithic baseline}
A single call receives a large system prompt (full registry, rules, and few-shot
exemplars) plus the rule, and emits workflow JSON. This mirrors the platform's
current authoring approach over templatized input, refined over many iterations,
in which few-shot exemplars are the dominant lever on accuracy. It is simple and,
for strong models, competitive; but it degrades for weaker generators on dense
rules---condition structure is incomplete and JSON validity suffers from
syntactic hallucination---and, as the registry and per-scenario exemplar set
grow, inlining all of it on every rule becomes increasingly costly.

\subsection{Intermediate representation and a target-agnostic compiler}
We split generation into \textbf{Step 1 (LLM)}: NL $\rightarrow$ \IR, a
low-syntax description of the workflow DAG, and \textbf{Step 2
(deterministic)}: \IR\ $\rightarrow$ JSON, a compiler that emits guaranteed-valid
workflow JSON. We instantiate the \IR\ as \textbf{YAML}---a deliberate choice, since its
light punctuation (no braces, few quotes, indentation-based nesting) asks the
model to emit far less structural syntax than the JSON target, directly reducing
the emission burden analyzed below; the method is agnostic to the exact
serialization. This split removes an entire class of failures: JSON validity
rises to 99--100\% and the compile step never fails on \IR\ output across any
model. Crucially, \textbf{Step 2 is deliberately agnostic to the specific end
workflow being generated}: the compiler translates the \IR\ into whatever target schema the
platform requires, so supporting a new workflow type or JSON dialect is a
compiler/registry change, not a prompt or model change. The compiler also
performs the rote expansions the model would otherwise hand-write---cross-products,
hit-first chaining, and list-valued condition operators that turn an
``all-except'' set into an AND of exclusions via De Morgan---so the model emits
each value list once instead of repeating it. The remaining gap after this split
is semantic---completeness and value precision---not syntax.

\subsection{Localizing the failure: the emission-density diagnostic}
Two measurements on the full benchmark (detailed in Section~\ref{sec:eval})
localize the failures to emission rather than interpretation. First, decomposing
accuracy by node granularity (Section~\ref{sec:headline}): the model almost
always selects the \textbf{right set of action nodes}---across the four configurations,
node-set accuracy ranges 98.4--99.8\% for GPT-5.2 and 99.5--100\% for Opus 4.6,
$\sim$93.5--97.6\% for GPT-4.1, and 82.4--92.6\% for GPT-5.3-chat---yet
node-exact accuracy (same nodes with correct attributes) is markedly lower (for
GPT-5.3-chat it ranges 62.5--82.8 across the same configurations). Read as
improvements over the monolithic baseline, \SelIR\ lifts GPT-5.3-chat's node-set
accuracy from 82.4 to 92.6 ($+10.2$ percentage points) and its node-exact
accuracy from 62.5 to 82.8 ($+20.3$ percentage points). The residual error is
therefore overwhelmingly wrong attributes on correctly-selected nodes, not wrong
selection. Second, stratifying every rule by branch-grid density
(Section~\ref{sec:density}, Figure~\ref{fig:density}) shows accuracy declining as
the number of interdependent nodes to emit at once grows, with monolithic
generation collapsing on dense rules while the compiler degrades gracefully. Both
point the same way: the model reads a rule well and drops content only as the
density of structure it must emit rises---ruling out ``help the model extract
better'' fixes and pointing at reducing how much structure the model must emit.

\subsection{The scaling front-end: learned registry selection}
A front-end call first \textbf{selects} the registry entries a rule needs (events,
actions, conditions, example IDs) as a small JSON; Step 1 then generates from a
prompt built only from that selection. Selection is accurate even with
non-reasoning models (selection F1 $\approx$ 0.96--0.99), and concise
disambiguation metadata in the registry resolves confusable condition pairs,
raising recall---the binding constraint, since generation can only emit what
selection retrieved.

\textbf{Why selection exists: scaling with registry size.} A monolithic prompt
inlines the entire vocabulary on every rule; as the registry grows beyond what
fits comfortably (or cheaply) in one prompt, this becomes untenable in cost and
accuracy. Selection keeps Step 1 focused on the handful of entries a rule
actually uses, independent of total vocabulary size. As the registry grows we
will front the LLM selector with a \textbf{lightweight similarity-based candidate
generator} (embedding retrieval over registry entries) that shortlists
likely-relevant entries before the LLM selects among them---keeping selection
high-recall and inexpensive, and bounding the selector's input regardless of
vocabulary size.

\subsection{Full system}
\SelIR\ feeds the filtered registry from the selector into the same \IR\ Step-1
prompt and compiles deterministically---combining the scaling front-end with the
validity-by-construction back-end.

\subsection{Why it works: a division of labor}
The unifying principle is a division of labor matched to where LLMs are strong
and weak:
\begin{itemize}
\item \textbf{LLMs are strong at language$\rightarrow$intent}: reading a rule and
identifying its segments, values, and intended actions (node-set accuracy
near-saturated, Section~\ref{sec:headline}).
\item \textbf{LLMs are weak at dense structured emission}: serializing many
interdependent nodes/edges at once, where they collapse the branch grid and
mis-wire boolean grouping.
\item \textbf{Compilers are exact at deterministic expansion}: cross-products, De
Morgan expansion of membership sets, hit-first chaining, residual handling.
\end{itemize}
Every remedy moves combinatorial emission to the compiler while leaving
interpretation with the LLM. The consequence, made precise below, is that the
benefit is \textbf{largest where monolithic prompting leaves the most headroom}: a model
that already emits dense structure well gains little from the compiler, while one
that under-produces it under monolithic prompting benefits substantially. This is
a transferable recipe for \NLgraph\ tasks on commodity models---one that narrows
rather than erases the gap to the strongest models.

\section{Evaluation}
\label{sec:eval}
\subsection{Setup}
\label{sec:setup}
Generation spans four models: \textbf{GPT-4.1} and \textbf{GPT-5.3-chat}
(non-reasoning), and \textbf{GPT-5.2} and \textbf{Claude Opus 4.6}
(reasoning-class references). GPT-4.1 is the
architecturally weakest model; GPT-5.3-chat has the most headroom under
monolithic prompting (it under-produces actions when handed the full monolithic
prompt). All 16 cells are generated \textbf{greedily (temperature 0)}, so decoding
variance is negligible and single-run uncertainty is quantified by resampling
rather than re-running. The semantic metric is an \textbf{LLM-as-judge
validity score (AI\%)} from Claude Opus 4.6 under one fixed rulebook, byte-identical across
configurations. Deterministic metrics---JSON validity, event/node-set match,
node-exact accuracy (correct action type and parameters), and exact
conditions-tree match (Cond\%)---are judge-independent: each is computed by
program directly against the ground-truth workflow for the rule, with no LLM in
the loop, so any result reported on Cond\% (including the density curve in
Section~\ref{sec:density}) stands or falls independently of the judge. Every cell
reports over all 635 rules.

\textbf{On judging Opus outputs with an Opus judge.} Self-preference bias is a
legitimate concern since one generator (Opus 4.6) shares the judge's family, but
it cannot drive our claims: our primary bridge involves no Opus-generated output
(GPT-5.3-chat vs.\ GPT-5.2 under the same judge, so any self-preference cancels),
and wherever an Opus generation is compared the bias is conservative---it
inflates the reasoning baselines our method must catch. The judge-independent
Cond\% metric preserves the same ordering. Appendix~\ref{app:bias} gives the full
three-part argument.

\begin{table}[t]
\centering\scriptsize
\resizebox{\columnwidth}{!}{%
\begin{tabular}{lcccc}
\toprule
Config & GPT-4.1 & GPT-5.3-chat & GPT-5.2 & Opus 4.6\\
 & \multicolumn{4}{c}{Cond\% / AI\%}\\
\midrule
Monolithic & 70.1 / 66.8 & 56.5 / 56.4 & 87.2 / 79.8 & 87.2 / 84.1\\
\SelJSON & 79.7 / 71.5 & 75.1 / 75.3 & 87.4 / 84.3 & 78.6 / 80.5\\
\IR-only & 78.3 / 65.2 & 79.4 / 77.2 & \textbf{91.0} / 82.4 & 83.0 / \textbf{88.8}\\
\SelIR & 78.1 / 64.6 & \textbf{82.2 / 80.6} & 89.6 / 84.1 & 88.0 / \textbf{88.8}\\
\bottomrule
\end{tabular}}
\sourcecaption{Exact-match condition accuracy (Cond\%) and LLM-judge validity
(AI\%) per configuration and model. \SelJSON\ uses a selector tuned independently
of \SelIR\ (it passes a fuller registry at selection time), so it is reported for
completeness and is not a single-variable ablation. Its token cost is in
Appendix~\ref{app:tokens}.}
\label{tab:headline}
\end{table}

\subsection{Headline results}
\label{sec:headline}
Exact-match condition accuracy (\textbf{Cond\%}) and LLM-judge validity
(\textbf{AI\%}) are in
Table~\ref{tab:headline}. JSON validity and event-type accuracy are 99--100\%
across all compiler cells (direct-emission configs slightly lower, e.g.\
Monolithic GPT-4.1 3.5\% invalid JSON) and omitted here.

The pipelines deliver a \textbf{large lift where monolithic prompting leaves the
most headroom}: GPT-5.3-chat's judge validity climbs 56.4 $\rightarrow$ 80.6 and
conditions 56.5 $\rightarrow$ 82.2. For \textbf{models that already emit dense
structure well the effect is muted}---monolithic Opus is already 84.1 AI\% / 87.2 Cond\%,
and the pipelines move it only a few points (and down on some cells).

Two further judge-independent metrics decompose node construction: \textbf{node-set}
accuracy (correct action types, ignoring parameters) and the stricter, primary
\textbf{node-exact} accuracy (correct types and parameters). Across configurations, node-set
accuracy ranges from 98.4--99.8\% for GPT-5.2, 99.5--100\% for Opus 4.6,
93.5--97.6\% for GPT-4.1, and 82.4--92.6\% for GPT-5.3-chat. For GPT-5.3-chat
specifically, \SelIR\ improves node-set accuracy from 82.4\% to 92.6\%
(+10.2 percentage points) and node-exact accuracy from 62.5\% to 82.8\%
(+20.3 points) relative to monolithic generation. So \textbf{most residual error
is parameter error on correctly-selected nodes, not wrong-node selection}---the model
knows which actions to emit but drops attributes as density grows, exactly the
emission-density bottleneck, confirmed without the LLM judge.
Appendix~\ref{app:nodes} gives the full per-model table.

\begin{figure*}[t]
\centering
\includegraphics[width=0.85\textwidth]{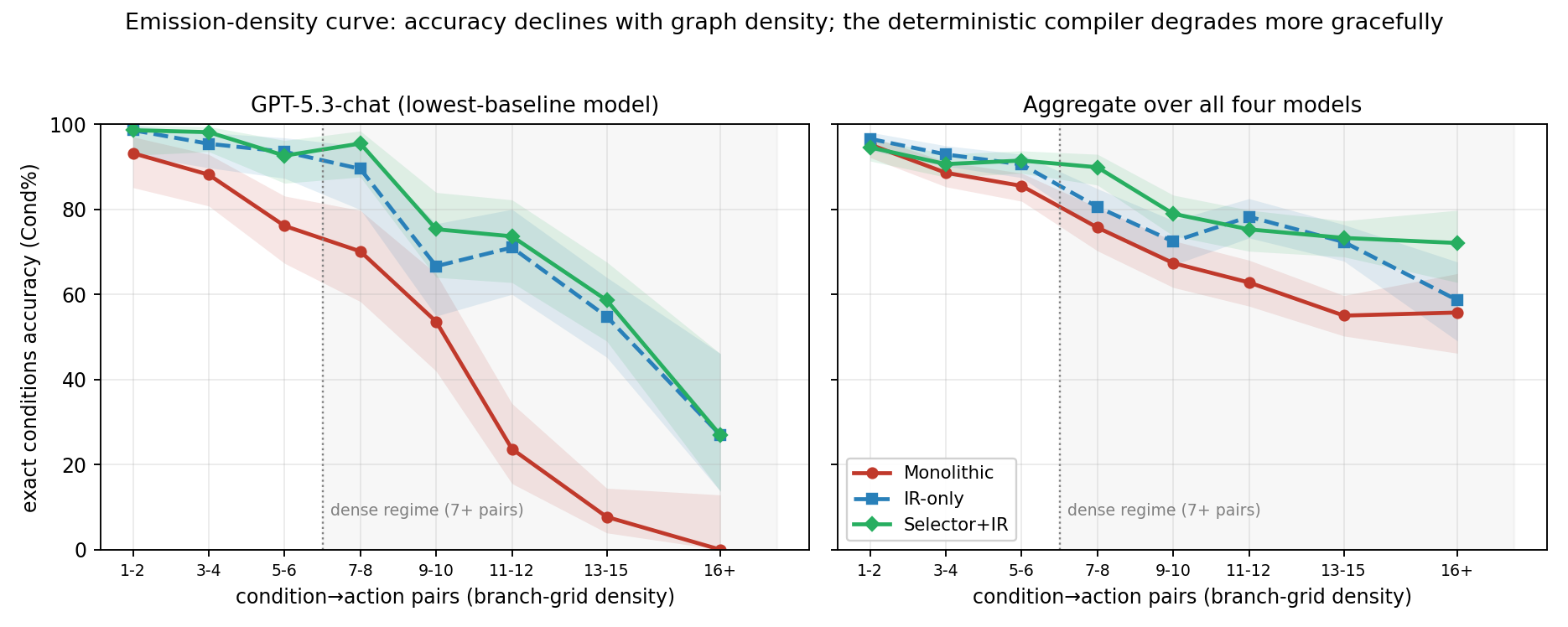}
\sourcecaption{Exact-conditions accuracy (corrected scorer) vs.\ branch-grid
density (condition$\rightarrow$action pairs). Left: GPT-5.3-chat; right:
aggregate over all four models. Shaded bands are Wilson 95\% CIs. Monolithic
generation collapses on dense rules while the compiler configurations degrade
gracefully---the visual signature of the emission-density bottleneck.}
\label{fig:density}
\end{figure*}

\subsection{The emission-density curve}
\label{sec:density}
Stratifying every rule by branch-grid density---the number of
condition$\rightarrow$action pairs, a proxy for how many interdependent nodes the
model must emit at once---reproduces the emission-density thesis on the full
benchmark (Figure~\ref{fig:density}). Every configuration declines as density
rises, but the compiler degrades far more gracefully. For GPT-5.3-chat the
contrast is stark: monolithic exact-conditions accuracy falls from $\sim$93\% at
1--2 pairs to $\sim$0\% beyond 15 pairs, whereas \SelIR\ still holds above
$\sim$55\% through the 13--15 band. Aggregated over all four models the ordering
is preserved. The shaded bands are Wilson 95\% confidence intervals on the
per-bin pass rate: the Wilson interval inverts the score test for a binomial
proportion, so unlike the normal (Wald) approximation it stays inside $[0,1]$ and
remains well calibrated when a density bin holds few rules or when the observed
accuracy sits near 0\% or 100\%---precisely the regime of the densest bands.

\subsection{Validity by construction}
The deterministic compiler guarantees structural validity:
\textbf{\IR-only produces zero uncompilable output across all four models}, with 99--100\% valid JSON and
$\sim$99\% correct event types throughout. Adding the selector introduces rare
malformed-\IR\ that almost always repairs, leaving exactly
\textbf{one hard failure across 5,080 pipeline generations (0.02\%)}---a disclosed robustness limitation of
selection, not a validity threat (per-model breakdown in
Appendix~\ref{app:compile}).

\subsection{Paired significance}
\label{sec:paired}
All contrasts are paired by rule ($n = 635$); because generation is greedy,
\textbf{McNemar's exact test} on per-rule pass/fail is the correct inferential tool. The
test conditions on the discordant rules---those a given pair of configurations
scores differently---and asks, under the null hypothesis that the two
configurations are equally accurate, whether the split of those disagreements
between the two directions is consistent with a fair coin; the $p$-value is the
exact binomial tail probability, so no large-sample approximation is required
even when the discordant count is small. Scenario-clustered bootstrap CIs show
that marginal per-cell rates carry $\pm15$--25-point uncertainty---so comparisons
must be read paired, not from overlapping marginal CIs. For \textbf{GPT-5.3-chat} every
pipeline beats monolithic by a wide, highly significant margin (\SelIR\ $+24.3$
points, $p<10^{-3}$)---the core, robust result. For \textbf{models that already
emit dense structure well} gains are small and metric-fragile: on exact conditions,
\SelJSON\ and \IR-only significantly reduce Opus's score (a scorer-strictness
effect on the compiler's canonical, logically-equivalent combiner nesting, which
the judge forgives). The full per-cell McNemar table is in
Appendix~\ref{app:sig}.

\subsection{Bridging non-reasoning and reasoning models}
Is ``non-reasoning models bridge the gap to reasoning models'' a fair claim? All
four models ran the same 635 rules, so cross-model comparison is paired. Because
the claim is one of similarity, we test it with two one-sided tests (TOST)
against a pre-specified, strict $\pm5$-point equivalence margin rather than
inferring parity from a non-significant difference test. The answer is
\textbf{yes, but tightly scoped. The bridge holds against GPT-5.2}:
GPT-5.3-chat + \SelIR\ (\textbf{80.6\% AI\%}) is statistically equivalent to
GPT-5.2's monolithic baseline (79.8\%;
paired difference $+0.8$pt, 90\% CI $[-2.0, +3.6]$, $p_{\mathrm{TOST}} = 0.007$).
\textbf{Against the stronger Opus 4.6 baseline (84.1\%) it reaches near-parity,
not formal equivalence}: the 3.5-point deficit is not individually significant
(McNemar $p = 0.067$), but equivalence holds only at a looser $\pm7$-point
margin. \textbf{Two further limits.} The architecturally weakest model (GPT-4.1) does not
bridge---even its best cell stays 8--13 points below both reasoning baselines
($p<10^{-4}$), and the same $\pm5$-point test that certifies GPT-5.3-chat rejects
GPT-4.1 (equivalence reached only at $\pm12$), evidence the procedure is
adequately powered at $n = 635$ and not merely rubber-stamping equivalence. And
the frontier gap persists---the best non-reasoning cell (80.6\%) trails the best
reasoning cell (Opus 4.6 + \IR-only, 88.8\%) by $-8.2$ points ($p<10^{-4}$),
because reasoning models also improve with the method. \textbf{Fair phrasing}:
deterministic structure lets a non-reasoning model match GPT-5.2's
monolithic-prompt quality and reach near-parity with Opus, narrowing but not
eliminating the gap; when the reasoning model also adopts the method, an
$\sim$8-point frontier gap remains. Full TOST results are in
Appendix~\ref{app:sig}.

\begin{table}[t]
\centering\scriptsize
\begin{tabular}{lrrrr}
\toprule
Config & GPT-4.1 & GPT-5.3 & GPT-5.2 & Opus 4.6\\
\midrule
Monolithic (1 call) & 23,246 & 21,264 & 23,767 & 27,286\\
\IR-only (1 call) & 20,122 & 19,311 & 20,323 & 23,906\\
 & (0.87$\times$) & (0.91$\times$) & (0.86$\times$) & (0.88$\times$)\\
\textbf{\SelIR} & \textbf{12,976} & \textbf{11,980} & \textbf{13,220} & \textbf{14,972}\\
\textbf{(2 calls)} & \textbf{(0.56$\times$)} & \textbf{(0.56$\times$)} & \textbf{(0.56$\times$)} & \textbf{(0.55$\times$)}\\
\bottomrule
\end{tabular}
\sourcecaption{Per-rule total tokens; ratio vs.\ monolithic in parentheses.}
\label{tab:tokens}
\end{table}

\subsection{Token efficiency}
\label{sec:tokeneff}
A multi-call pipeline could plausibly cost more than one monolithic call; it does
not. We report per-rule tokens (objective and provider-independent; they
translate to cost under any pricing) in Table~\ref{tab:tokens}.

The saving has \textbf{two independent sources}. First, \textbf{\IR\ alone} helps
($\sim$0.87$\times$): emitting compact \IR\ instead of verbose JSON shrinks the
completion (e.g., 4.1K $\rightarrow$ 2.5K tokens on GPT-4.1). Second,
\textbf{the selector trims the prompt}: monolithic inlines the entire verbose registry (the
bulk of its $\sim$19K-token prompt) on every rule, whereas the selector call
carries only a compact $\sim$2.2K menu and the assembler only the selected
subset---taking \SelIR\ to $\sim$0.56$\times$ on every model. The bridging cell is
thus both competitive and cheaper: GPT-5.3-chat + \SelIR\ matches the reasoning
models' monolithic quality using 0.44--0.50$\times$ the tokens of their
monolithic runs.

\subsection{Error analysis}
The residual errors are concentrated and genuine. Within the recurring-timer
family, $\sim$40\% of cells still miss exact conditions---real model errors
(wrong eligibility trees, dropped phases) corroborated by lower judge scores, not
syntax or scoring artifacts---and they define the open challenge.

\subsection{What structure fixes, and what it does not}
\label{sec:whatfixes}
To characterize how the pipeline helps, we automatically categorized the judge's
rationale on all 2,348 failing cells (aggregate trends only; method and
per-category numbers in Appendix~\ref{app:taxonomy}). The errors fall into two
classes with sharply different responses to decomposition
(Figure~\ref{fig:errors}, Appendix~\ref{app:taxonomy}).

\textbf{Completeness and syntax errors collapse}: the pipelines roughly halve
incomplete-rule and De Morgan/exclusion errors and all but eliminate raw
JSON-structure errors---this accounts for most of the AI-quality gain.
\textbf{Logic and topology errors persist}: cross-variable AND/OR flips stay
flat, and ordering errors---the single largest category (49.5\% of failures),
i.e.\ wrong edge direction in the DAG---barely move, with only \SelIR\ reducing
them meaningfully. Externalizing emission thus converts omission errors into
mis-ordering errors, leaving DAG topology as the dominant residual. The mix is
model-dependent: the lowest-baseline model's failures are dominated by
incompleteness (76\%) while the strongest model's are almost entirely
boolean-logic errors. This reads the judge's rationales rather than ground truth,
but the judge-independent conditions scorer confirms the persistence of
boolean-structure errors.

\section{Deployment Status and Path}
\label{sec:deploy}
\textbf{Status.}
The platform's current authoring capability generates workflows from a
\textbf{templatized} rule representation using a monolithic prompt. The system here
extends that along two axes the templatized approach does not
address---accepting \textbf{text rules} and \textbf{scaling to a growing registry} without an
ever-expanding prompt---and is validated in offline evaluation at scale (635
rules, four models). Its architecture was chosen for production cost, targeting
the same operating envelope as the current system.

\textbf{Path toward scale.}
New events/actions/conditions are added as registry entries with disambiguation
metadata; the compiler and prompts are unchanged, and few-shot exemplars are
single-sourced from the registry. The learned selector is the front-end for
registry growth, and a \textbf{similarity-based candidate prefilter} (embedding retrieval
before the LLM selector) keeps Step 1 high-recall and inexpensive as the
vocabulary expands. Because \textbf{Step 2 is agnostic to the target workflow}, extending
to new schemas is a compiler/registry change, not a model change.

\textbf{Validating the judge.}
Our headline semantic metric is an LLM-as-judge score;
Section~\ref{sec:setup} argues self-preference bias cannot drive our central
claims and in fact strengthens them (Appendix~\ref{app:bias}), and the
judge-independent conditions metric agrees throughout. To check the judge against
human reading, two authors labeled a verdict-balanced (50 judge-PASS, 50
judge-FAIL), stratified 100-item sample \textbf{blind to the judge verdict, the
deterministic score, and the generating model and method}. The two labeling
assignments overlapped only partially: one author labeled 82 items and the other
40, with 22 items labeled independently by both. Agreement with the judge was
\textbf{91\% (combined Cohen's $\kappa = 0.82$)} and inter-annotator agreement on the
22-item overlap was $\kappa = 0.82$. As both annotators are authors, this is a
blind consistency check; blindness rules out anchoring, and the conditions metric
supplies independent corroboration (Appendix~\ref{app:human}).

\textbf{Generality.}
The architecture---LLM $\rightarrow$ \IR\ $\rightarrow$ deterministic compiler
with learned selection and a similarity prefilter---is domain-agnostic; the
emission-density diagnostic and evaluation practices should transfer to other
\NLgraph\ tasks (business-process models, ETL/dataflow graphs, agent tool-graphs,
CI/build pipelines).

\section{Lessons Learned}
\label{sec:lessons}
We surface two transferable lessons that materially affect whether such a system
is measured correctly; two further practices (treating the scorer as fallible
code, and how greedy generation buys honest paired statistics) are in
Appendix~\ref{app:lessons}.

\textbf{Align the judge's rulebook with the generator's.}
Convention drift between generator and judge is a first-order confound: a judge
that reads a surface ``or'' literally will false-fail a generator that correctly
used De Morgan. We pin one rulebook for every system compared and treat any
configuration that looks anomalously bad as suspect until re-checked.

\textbf{Decomposition relocates the bottleneck; it does not remove it.}
Splitting generation into plan$\rightarrow$expand or code-locked scaffolds
consistently failed---the node-dropping collapse simply moved to the
content-filling step. Reduce what must be emitted (\IR, list-valued conditions,
selection); do not merely chop the generation step.

\section{Conclusion}
Generating workflow DAGs from natural language is a real enterprise need, and it
can be met \textbf{without the cost of extended-reasoning models}. Commodity LLMs fail at
emitting dense structure, not at understanding it; moving combinatorial graph
construction into a deterministic compiler---via an \IR\ and a learned,
similarity-prefiltered registry selector---lets non-reasoning models reach
production-relevant operating points with guaranteed structural validity at
roughly half the token cost. Our sharpest, honestly scoped result is a
gap-bridging effect: structure delivers a large, significant lift where
monolithic prompting leaves the most headroom, lifting a non-reasoning model to
statistically match (by an equivalence test) a reasoning model's out-of-the-box
quality, while an $\sim$8-point frontier gap persists. We hope the
emission-density framing, the deployment path, and the evaluation practices help
other teams ship and honestly measure structured-generation applications on
affordable models.

\section*{Data Availability}
The evaluation data and the platform's vocabulary are proprietary and governed by
confidentiality agreements; we report aggregate metrics and dataset
characteristics (size, scenario-family structure, condition-density distribution)
rather than releasing examples or schemas. The methods are described in enough
detail to reimplement on any comparable structured-workflow target.

\appendix
\section{Self-preference bias --- full argument (supports Sec.~\ref{sec:setup})}
\label{app:bias}
Because one of the four generators (Opus 4.6) shares a model family with the
judge, self-preference bias---an LLM judge rating its own outputs more
favorably---is a legitimate concern. Three observations bound its impact on our
conclusions. \textbf{First, our primary bridging comparison involves no
Opus-generated output}: GPT-5.3-chat + \SelIR\ (80.6\% AI\%) versus GPT-5.2
monolithic (79.8\%) pits two non-Opus generations against the same Opus judge, so
any self-preference cancels. \textbf{Second, wherever the comparison does include
an Opus generation, self-preference is not merely harmless but actively
strengthens our case}: it inflates Opus's monolithic baseline (making the
``non-reasoning model approaches Opus'' bridge harder to claim) and inflates the
best reasoning cell (widening the reported $\sim$8-point frontier gap), so the
true gap our method must close is, if anything, smaller than reported. The bias
thus works against every claim we make---it can only understate the
non-reasoning story, never manufacture it. Any residual self-preference makes our
reported lifts a conservative lower bound. \textbf{Third, the judge-independent
Cond\% metric preserves the same qualitative ordering} (e.g., Opus monolithic
87.2 vs.\ GPT-5.3-chat + \SelIR\ 82.2), so the judge is not fabricating the
picture. A blind, author-labeled human-agreement study corroborates the judge
(91\% agreement, $\kappa = 0.82$; Appendix~\ref{app:human}).

\section{Node-set / node-exact accuracy, per model (supports
Sec.~\ref{sec:headline})}
\label{app:nodes}
Node-set accuracy scores correct action types (ignoring parameters); node-exact
accuracy is the stricter metric requiring correct types and parameters (exact
key/value match).

\begin{table}[t]
\centering\scriptsize
\resizebox{\columnwidth}{!}{%
\begin{tabular}{lcccc}
\toprule
Config & GPT-4.1 & GPT-5.3 & GPT-5.2 & Opus 4.6\\
 & \multicolumn{4}{c}{Node-set / Node-exact}\\
\midrule
Monolithic & 95.7 / 80.3 & 82.4 / 62.5 & 98.4 / 86.1 & 99.5 / 89.8\\
\SelJSON & 97.6 / 83.5 & 89.3 / 80.2 & 99.8 / 89.3 & 100 / 89.3\\
\IR-only & 93.5 / 76.5 & 90.2 / 81.3 & 99.2 / 89.1 & 100 / 90.7\\
\SelIR & 94.6 / 77.6 & \textbf{92.6 / 82.8} & 99.2 / 88.5 & 99.8 / \textbf{91.7}\\
\bottomrule
\end{tabular}}
\sourcecaption{Node-set and node-exact accuracy per configuration and model.}
\label{tab:nodes}
\end{table}

The decomposition localizes the residual error: node-set is 98--100\% for GPT-5.2
and Opus 4.6 ($\sim$93--98\% for GPT-4.1) and moves only modestly for
GPT-5.3-chat (82.4 $\rightarrow$ 92.6), so most of the remaining error is
parameter error on correctly-selected nodes.

\section{Compile-failure breakdown (supports Sec.~\ref{sec:paired})}
\label{app:compile}
\IR-only produces zero uncompilable output across all four models. Adding the
selector introduces rare malformed-\IR\ (compile-failure rate: GPT-4.1 2.0\%,
Opus 4.6 0.2\%, GPT-5.3-chat and GPT-5.2 0\%); all but one of these repair to
valid JSON, leaving exactly one hard failure across 5,080 pipeline generations
(0.02\%).

\begin{figure*}[t]
\centering
\includegraphics[width=0.80\textwidth]{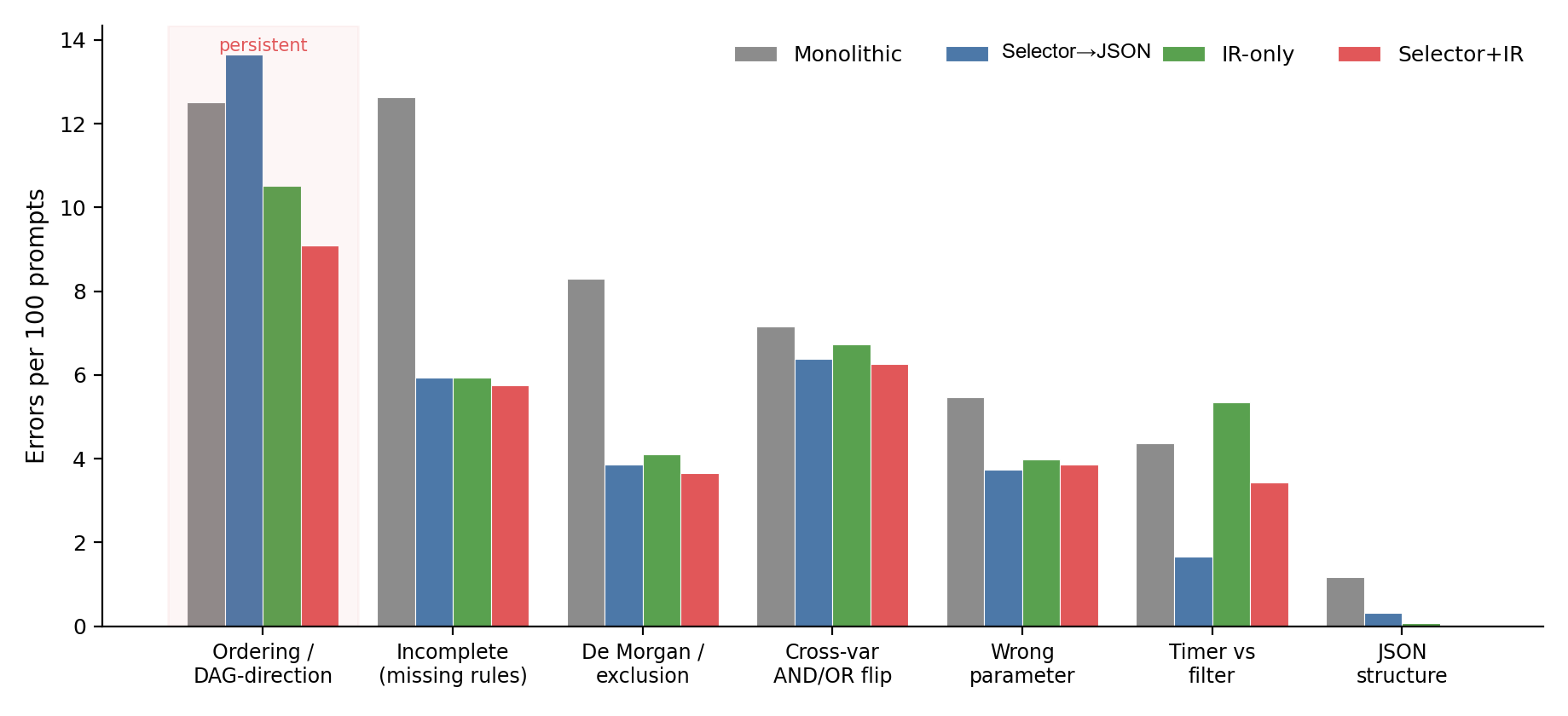}
\sourcecaption{Error prevalence (per 100 prompts) by category and configuration,
from automated coding of the judge's rationales. Completeness/syntax categories
(Incomplete, De Morgan, JSON) collapse under the pipelines; logic/topology
categories (cross-variable AND/OR, Ordering) persist. Ordering---wrong DAG edge
direction---is the largest and most stubborn residual.}
\label{fig:errors}
\end{figure*}

\section{Full paired-significance table (supports Sec.~\ref{sec:paired})}
\label{app:sig}
All contrasts are paired by rule ($n = 635$). We report scenario-clustered
bootstrap 95\% CIs (resampling the seven scenario families, then rules within),
which show that marginal per-cell rates carry $\pm15$--25-point uncertainty.

Each configuration vs.\ Monolithic (judge validity; $\Delta$ points, McNemar $p$;
*** $p<10^{-3}$, * $p<0.05$, ns not significant):

\begin{table}[t]
\centering\scriptsize
\begin{tabular}{lccc}
\toprule
Model & \SelJSON & \IR-only & \SelIR\\
\midrule
GPT-4.1 & $+4.7$ (*) & $-1.6$ (ns) & $-2.2$ (ns)\\
GPT-5.3-chat & $+18.9$ (***) & $+20.8$ (***) & $+24.3$ (***)\\
GPT-5.2 & $+4.4$ (*) & $+2.5$ (ns) & $+4.3$ (*)\\
Opus 4.6 & $-3.6$ (ns) & $+4.7$ (*) & $+4.7$ (*)\\
\bottomrule
\end{tabular}
\sourcecaption{Judge-validity deltas vs.\ Monolithic, with McNemar
significance.}
\label{tab:mcnemar}
\end{table}

On exact conditions, \SelJSON\ and \IR-only significantly reduce Opus's score:
the compiler's canonical combiner nesting differs from ground-truth structure
though it is logically equivalent---a scorer-strictness effect the judge
forgives.

\subsection*{Equivalence tests (TOST) for the bridging claim}
Because the bridge is a claim of similarity, not difference, we test it with two
one-sided tests (TOST) against a pre-specified $\pm5$-point equivalence margin (a
strict operating tolerance). We report the paired mean difference, its 90\% CI,
$p_{\mathrm{TOST}}$ at $\pm5$, and the smallest margin at which equivalence holds
($p_{\mathrm{TOST}}<0.05$). All tests are paired by rule ($n = 635$) on per-rule
judge pass/fail.

\begin{table*}[t]
\centering\scriptsize
\begin{tabular}{lrrrr}
\toprule
Comparison (AI\%) & $\Delta$pt & 90\% CI & $p_{\mathrm{TOST}}^{\pm5}$ &
smallest equiv.\ margin\\
\midrule
GPT-5.3 \SelIR\ vs GPT-5.2 mono & $+0.8$ & $[-2.0, +3.6]$ & 0.007 & $\pm4$pt\\
GPT-5.3 \SelIR\ vs Opus 4.6 mono & $-3.5$ & $[-6.4, -0.5]$ & 0.198 & $\pm7$pt\\
GPT-4.1 best vs GPT-5.2 mono & $-8.4$ & $[-11.8, -4.9]$ & --- & $\pm12$pt\\
GPT-4.1 best vs Opus 4.6 mono & $-12.6$ & $[-15.9, -9.3]$ & --- & $>\pm15$pt\\
\bottomrule
\end{tabular}
\sourcecaption{TOST equivalence tests for the bridge. GPT-4.1 ``best'' is its
highest-AI\% cell (\SelJSON, 71.5\%). Equivalence to GPT-5.2 holds at the strict
$\pm5$-point margin; equivalence to Opus 4.6 holds only at $\pm7$ (reported as
near-parity); GPT-4.1 is rejected at $\pm5$.}
\label{tab:tost}
\end{table*}

The procedure has \textbf{discriminant validity}: at the same $n$ and the same
$\pm5$-point margin it certifies GPT-5.3-chat + \SelIR\ as equivalent to GPT-5.2
yet rejects equivalence for GPT-4.1's best cell (which needs $\pm12$), so a
positive equivalence verdict reflects genuine parity rather than low power.
Equivalence to Opus 4.6 is not established at $\pm5$ (only at $\pm7$), so we
report near-parity there, not a match.

\textbf{Single-run scope (threat to validity).} Every cell is a single greedy
(temperature-0) generation, so these intervals capture rule-sampling and scorer
uncertainty but not generation stochasticity, prompt-paraphrase sensitivity, or
few-shot-ordering sensitivity. The qualitative ordering is reproduced by the
judge-independent Cond\% metric, but quantifying decode- and prompt-level
variance is left to future work and is the main residual threat to the point
estimates.

\section{Error-taxonomy coding method and per-category numbers (supports
Sec.~\ref{sec:whatfixes})}
\label{app:taxonomy}
We automatically categorized the judge's text rationale on all 2,348 failing
cells across the four configurations and four models by keyword coding gated on
negative cues (96\% coverage; aggregate trends only). Relative to the monolithic
baseline, the pipelines roughly halve missing-rule (incomplete) errors
(12.6 $\rightarrow$ 5.7 per 100 prompts) and De Morgan / exclusion errors
(8.3 $\rightarrow$ 3.7), and all but eliminate raw JSON-structure errors
(1.2 $\rightarrow$ 0.0). Cross-variable AND/OR connective flips stay essentially
flat ($\sim$6.3--7.2 across all configurations). Ordering errors---the single
largest category (49.5\% of all failures)---barely move and even worsen under
\SelJSON\ (12.5 $\rightarrow$ 13.7); only \SelIR\ reduces them meaningfully
($\rightarrow$ 9.1). This analysis is corroborated by the blind human-agreement
study (Appendix~\ref{app:human}) and by the judge-independent conditions scorer.

\section{Two further evaluation lessons (supports Sec.~\ref{sec:lessons})}
\label{app:lessons}
\textbf{Treat the deterministic scorer as code that can be wrong.} A metric that
is uniformly degenerate across many independent models and configurations is
almost always a scorer bug, not a result. Our conditions scorer resolves gates
inherited transitively down timer cascades (walking dependency edges upward
through sequencing nodes, but only across success edges so fallback branches do
not inherit a gate); we validate it against the judge and unit tests.

\textbf{Greedy generation buys cheap, honest statistics.} Because every cell is
temperature-0, paired McNemar on a single run is the correct test;
scenario-clustered bootstrap CIs then honestly expose that marginal per-cell
rates are far less certain than a na\"ive binomial CI implies.

\section{Full per-config token breakdown (supports Sec.~\ref{sec:tokeneff})}
\label{app:tokens}
The main-text token table (Section~\ref{sec:tokeneff}) reports the three clean
cost configurations (Monolithic, \IR-only, \SelIR). For completeness we include
the selection-only variant (\SelJSON), which emits JSON directly after selection.
It is not cheaper than the monolithic baseline
($\approx$1.08--1.12$\times$)---not because it re-inlines the full registry (it
does not; its JSON call carries only the selected registry), but because it pays
two full-size calls: a selection call ($\sim$9.3K tokens on GPT-4.1, which reads
a substantial system prompt) followed by a JSON-generation call over the selected
registry ($\sim$15.8K tokens: $\sim$13.2K prompt + $\sim$2.6K completion). The
selection overhead exceeds the prompt saved by trimming the registry, so the
two-call total exceeds a single monolithic call. Because this variant's selector
was tuned independently of the \SelIR\ selector, its numbers are not a clean
single-variable ablation.

\begin{table}[t]
\centering\scriptsize
\begin{tabular}{lrrrr}
\toprule
Config & GPT-4.1 & GPT-5.3 & GPT-5.2 & Opus 4.6\\
\midrule
Monolithic (1 call) & 23,246 & 21,264 & 23,767 & 27,286\\
\SelJSON & 25,117 & 23,846 & 26,052 & 30,688\\
(2 calls) & (1.08$\times$) & (1.12$\times$) & (1.10$\times$) & (1.12$\times$)\\
\IR-only (1 call) & 20,122 & 19,311 & 20,323 & 23,906\\
 & (0.87$\times$) & (0.91$\times$) & (0.86$\times$) & (0.88$\times$)\\
\textbf{\SelIR} & \textbf{12,976} & \textbf{11,980} & \textbf{13,220} & \textbf{14,972}\\
\textbf{(2 calls)} & \textbf{(0.56$\times$)} & \textbf{(0.56$\times$)} & \textbf{(0.56$\times$)} & \textbf{(0.55$\times$)}\\
\bottomrule
\end{tabular}
\sourcecaption{Per-rule total tokens; ratio vs.\ monolithic in parentheses.}
\label{tab:fulltokens}
\end{table}

\section{Blind human-agreement study (supports Sec.~\ref{sec:setup},
Sec.~\ref{sec:deploy})}
\label{app:human}
Two authors labeled a 100-item sample as valid/invalid, blind to the judge
verdict, the deterministic score, the judge's rationale, and the generating model
and method. The two assignments overlapped only partially: Annotator 1 labeled 82
items, Annotator 2 labeled 40, and 22 items were labeled independently by both.
The sample is verdict-balanced (50 judge-PASS, 50 judge-FAIL) to neutralize the
prevalence effect on $\kappa$, and stratified across judging group (opus
self-judge 40, cross-model 60), method (baseline/\IR-only/\SelJSON/\SelIR, 25
each), generating model (Opus 40; GPT-4.1/GPT-5.2/GPT-5.3-chat 20 each), and
branch-grid density (1--2: 12, 3--5: 27, 6--9: 24, 10+: 37). Y/N labels map to
PASS/FAIL and are compared against the judge verdict.

\begin{table}[t]
\centering\scriptsize
\begin{tabular}{lrrr}
\toprule
Comparison & $n$ & Raw agreement & Cohen's $\kappa$\\
\midrule
Annotator 1 vs.\ judge & 82 & 93.9\% & 0.878\\
Annotator 2 vs.\ judge & 40 & 90.0\% & 0.802\\
Inter-annotator (overlap) & 22 & 90.9\% & 0.818\\
\textbf{Combined (40 + 60)} & \textbf{100} & \textbf{91.0\%} & \textbf{0.820}\\
\bottomrule
\end{tabular}
\sourcecaption{Blind human-vs-judge agreement. The combined row uses one label
per row---all 40 rows Annotator 2 labeled plus the 60 remaining
(Annotator-1-only) rows---for full 100-item coverage.}
\label{tab:human}
\end{table}

All comparisons fall in the ``substantial'' to ``almost perfect'' agreement band,
indicating the judge tracks human reading under blind conditions.

\bibliography{Iyer}

\end{document}